\documentclass[letterpaper, 10 pt, conference]{ieeeconf}  
\usepackage{graphicx}

\IEEEoverridecommandlockouts                              
\usepackage{amsmath,amssymb,bbm}
\usepackage{xcolor}
\usepackage{optidef}
\usepackage[normalem]{ulem}
\usepackage{lipsum}
\usepackage{mathtools}
\usepackage{cuted}
\usepackage{algorithm}
\usepackage{algpseudocode}
\usepackage{breqn}
\usepackage{subfig}
\usepackage{psfrag}
\usepackage{graphicx}
\usepackage{hyperref}
\usepackage{booktabs}
\usepackage{cite}
\usepackage{svg}
\usepackage[T1]{fontenc}  
\usepackage[utf8]{inputenc}

\newtheorem{theorem}{\hspace{0pt}\bf Theorem}
\newtheorem{assumption}{\hspace{0pt}\bf Assumption}
\newtheorem{lemma}{\hspace{0pt}\bf Lemma}
\newtheorem{proposition}{\hspace{0pt}\bf Proposition}

\newcommand\redsout{\bgroup\markoverwith{\textcolor{red}{\rule[0.5ex]{2pt}{0.4pt}}}\ULon}

\definecolor{mygreen}{rgb}{0.10,0.50,0.10}

\usepackage{graphicx}

\title{\LARGE \bf Transfer Learning for a Class of Cascade Dynamical Systems
}

\author{Shima Rabiei$^\dagger$, Sandipan Mishra$^\star$ and Santiago Paternain$^\dagger$
\thanks{$^\dagger$The authors are with the Department of Electrical, Computer, and Systems Engineering, Rensselaer Polytechnic Institute. Email: \{rabiei, paters\}@rpi.edu 
\newline
\indent $^\star$ The author is with the Department of Mechanical, Aerospace and Nuclear Engineering, Rensselaer Polytechnic Institute. Email: msihr2@rpi.edu 
}}

\begin{document}

\maketitle
\thispagestyle{empty}
\pagestyle{empty}

\begin{abstract}
This work considers the problem of transfer learning in the
context of reinforcement learning. Specifically, we
consider training a policy in a reduced order system
and deploying it in the full state system. The motivation
for this training strategy is that running simulations in
the full-state system may take excessive time
if the dynamics are complex. While transfer learning
alleviates the computational issue, the transfer guarantees
depend on the discrepancy between the two systems.
In this work, we consider a class of cascade dynamical
systems, where the dynamics of a subset of the state-space
influence the rest of the states but not vice-versa. The
reinforcement learning policy learns in a model that
ignores the dynamics of these states and treats them as
commanded inputs. In the full-state system, these dynamics
are handled using a classic controller (e.g., a PID). These
systems have vast applications in the control literature
and their structure allows us to provide transfer
guarantees that depend on the stability of the inner loop
controller. Numerical experiments on a quadrotor support
the theoretical findings.

\end{abstract}

\section{Introduction}

Reinforcement Learning (RL) has successfully solved control problems with complex dynamics and uncertainty. Its applications include robotics~\cite{levine2016end, garaffa2021reinforcement}, power systems~\cite{ernst2004power,dwivedi2024blackout} and aerial vehicles~\cite{abbeel2006application,liu2022deep} among others. 

Their success notwithstanding, these algorithms require large amounts of data to train and excessive time to converge~\cite{arulkumaran2017deep,kamthe2018data}. Moreover, the complexity of these problems 
increases with the dimensionality of the state-action space, something known as the curse of dimensionality~\cite{bellman1966dynamic}. To alleviate this burden (and also due to safety considerations), it is common to train the RL policies in simulation~\cite{osinski2020simulation,jayarathne2023safe}. Simulators, however, are a double edge sword. On the one hand, high fidelity simulations consider the full state of the system and the time to run them could be orders of magnitude larger than the process that is being simulated~\cite{muratore2021data,jayarathne2023safe}. Although this reduces the amount of data needed from the real system, it does not solve the problem of RL algorithms taking excessive time to converge. Hence, to improve the running time, it is not uncommon to consider reduced-order models with simplified dynamics~\cite{jayarathne2023safe,chen2023,chen2024invertedpendulumstask,Kajita1991StudyOD,Furusho1987ATM}. However, these suffer from the sim2real gap (see e.g., ~\cite{zhao2020sim}). Specifically, a policy that has been trained for certain dynamics may not necessarily maintain its performance when deployed in the real world (or a different environment).

Several approaches have been proposed in the literature to bridge this gap. System identification, parameter estimation or model learning focus on calibrating simulation models to match real-world dynamics using collected data~\cite{koos2013transferability}. Another line of work consists in fine-tuning the policy learned in simulation with real-world data~\cite{zhang2015vision,deisenroth2011pilco,rusu2016sim}. Meta-learning follows this idea, however, during the training stage agents are presented with several multiple task. This enables agents to quickly adapt to new environments, facilitating adjustment to differences between simulation and reality ~\cite{finn2017model,tobin2017domain}. A related approach is adversarial training, where disturbances are introduced to improve robustness to unexpected scenarios ~\cite{pinto2017robust}. Curriculum learning trains agents progressively with increasing complexity to aid generalization~\cite{bengio2009curriculum}. 

In this work, we take the approach of exploiting the structure of the dynamical system to guarantee transferability. In particular, we focus in a class of cascade dynamical systems, where a subset of the state influences the rest but not vice-versa. These problems are motivated by aerial vehicles, where the linear accelerations depend on their attitude~\cite{beard2012quadrotor,jayarathne2023safe}, but they find applications in other domains such as robotics~\cite{spong1987model} or control of temperature, flow and pressure~\cite{kaya2007improved}. In these scenarios, it is not uncommon to design controllers with nested loops. Often, the inner loops have simple dynamics and classic controllers, e.g., PID, can be designed~\cite{ogata2010modern}. Our approach is then, to train the RL policies in reduced order models (where these inner states are treated as inputs), and then deploy them in the full state system (See Figure \ref{fig:diagram}). 

In the next section, we formalize the transfer learning problem and the hypotheses that define the class of dynamical systems considered. Section \ref{sec_main_result} provides the main results of this work, where we establish bounds in performance degradation under input-to-state stability of the inner loop dynamics. Other than concluding remarks, this work finishes with numerical experiments (Section \ref{sec_numerical_examples}) where we validate the theoretical findings in a quadrotor navigation problem.

%
%
%
%
%
%

\section{Problem Statement}\label{sec_problem_formulation}
\begin{figure}
    \centering
    \includegraphics[width=\linewidth]{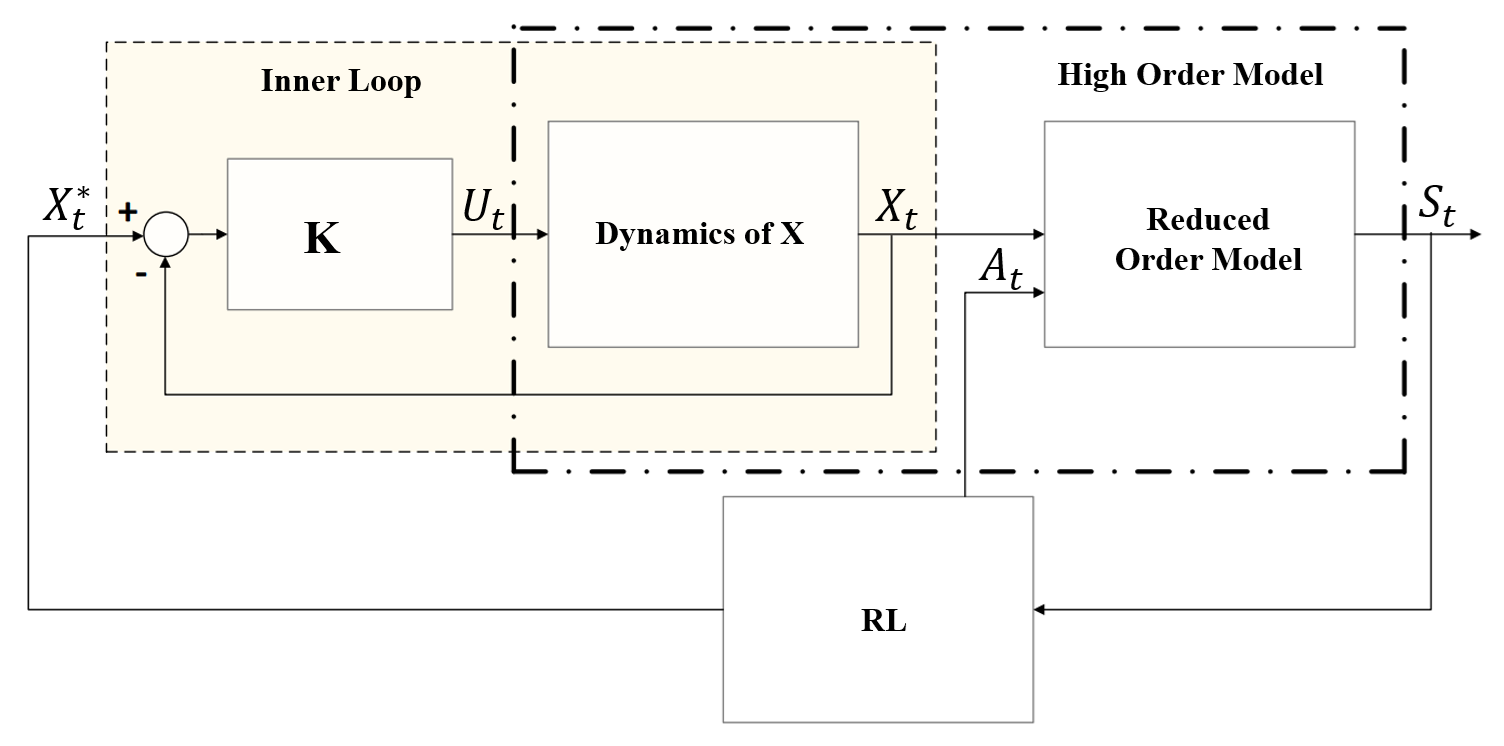} 
    \caption{Overview of the proposed approach. The RL agent is trained only on the reduced order model, where the inner loop is considered to have a unit transfer function. Thus, $X_t^\star$ and $A_t$ are the actions of the RL agent. The learned policy is then transferred to the full system that includes the dynamics of the state $X$. }
    \label{fig:diagram}
\end{figure}

In this paper, we consider the problem of training a policy in a reduced order model and transfer it to the full state dynamical system. To be formal, let us denote by \(\mathcal{S} \times \mathcal{X} \in \mathbb{R}^n\times\mathbb{R}^m\) the state space of the full state dynamical system and let $\mathcal{A}\times \mathcal{U} \in \mathbb{R}^p\times \mathbb{R}^q$ be its action space. The reduced order model is such that its state space is only $\mathcal{S}$ and its actions space is $\mathcal{A}\times \mathcal{X}$ (see Figure \ref{fig:diagram}). The dynamics of the systems are characterized by the state transitions probabilities. Let $\Delta(\cdot)$ represent the simplex over a set and denote by $\mathbb{P}_{R}: \mathcal{S} \times (\mathcal{A} \times \mathcal{X} ) \to \Delta (\mathcal{S}$), and \ $\mathbb{P}_{H} : (\mathcal{S} \times \mathcal{X}) \times (\mathcal{A} \times \mathcal{U}) \to \Delta(\mathcal{S}\times \mathcal{X})$ the transition probabilities of the reduced order model and the full state system respectively. These represent the probabilities to transition to a state given the current state and the action selected.
%
%
We consider scenarios where the optimal policy is trained on the reduced-order model and deployed on the full-state system. To do so, consider the reward function $r:\mathcal{S}\times\mathcal{A}\times\mathcal{X}$ and define the optimal policy as
\begin{equation}\label{eqn_optimal_reduced}
\pi_{R}^\star = \arg\max_{\pi} \mathbb{E}_{R}\left[ \sum_{t=0}^{\infty} \gamma^{t} r(S_t, A_t, X_t) \right],
\end{equation}
where the expectation is defined with respect to the transition probabilities of the reduced-order MDP as well as the actions which could be provided by a stochastic policy. {To ensure finite expected returns, a requirement for convergence of many algorithms in RL (see e.g., ~\cite{tsitsiklis1994asynchronous,sutton2018reinforcement}), we assume bounded rewards. } We formalize this assumption next.

\begin{assumption}\label{assump:reward_func}
 The rewards are bounded by {a constant}  \( B > 0 \), i.e.,
$|r(s, a,x)| \leq B$ {for all}  $(s, a,x) \in \mathcal{S} \times \mathcal{A}\times \mathcal{X}$.
\end{assumption}

The problem above can be solved through a myriad of methods,  e.g.,~\cite{tsitsiklis1994asynchronous,sutton1999policy}. Since the solution depends on the dynamics of the reduced-order model (which ignore the dynamics of $X$), its performance when transferred to the full state system, may degrade. Moreover, the policy cannot be directly applied to the high-order system as it does not provide an action $U_t\in\mathcal{U}$. Instead, the output of the RL policy can be considered a reference $X_t^\star$ for state $X_t$ (see Figure \ref{fig:diagram}). Then, a controller (depicted with the block K in Figure \ref{fig:diagram}) is in charge tracking the reference by setting 
\begin{equation}
U_t = K(X_t^\star, X_t).
\end{equation}
{In what follows we use the subindex $K$ to represent the MDP that results from incorporating the inner controller to the high-order MDP. We define the expected cumulative reward}
under the transferred policy $\pi_{R}^\star$ as 
\begin{equation}\label{eqn_high_order_controller_value}
V_{K}^{\pi_{R}^\star} = \mathbb{E}_{K}\left[ \sum_{t=0}^{\infty} \gamma^{t}r(S_t, A_t, X_t,X_t^{\star}) \mid  \pi_{R}^\star\right]. 
\end{equation}
The above quantity is our primary metric for evaluating the performance degradation of transferring the optimal policy $\pi^\star_R$ in \eqref{eqn_optimal_reduced}.  In Section \ref{sec_main_result} we provide guarantees on this degradation. Before doing so, we formalize and discuss the assumptions on these systems.

%
\begin{assumption}\label{assump:x_dynamics}
The dynamics of the state $X_t\in\mathcal{X}$ are independent of the state $S_t\in\mathcal{S}$ and action $A_t\in\mathcal{A}$, i.e., 
\begin{equation}
\mathbb{P}_{H}(X_{t+1} \mid X_{t}, S_{t}, A_{t}, U_{t}) = \mathbb{P}_{H}(X_{t+1} \mid X_{t}, U_{t}).
\end{equation}
\end{assumption}

\begin{assumption}\label{assump:s_dynamics}
For all $U_{t} \in \mathcal{U}$ it holds that:
\begin{equation}
\mathbb{P}_{H}(S_{t+1} \mid S_{t}, A_{t}, X_{t}, U_{t})  = \mathbb{P}_{R}(S_{t+1} \mid S_{t}, A_{t},X_{t}).
\end{equation}
\end{assumption}
Assumption \ref{assump:x_dynamics} states that the state $X_{t+1}\in\mathcal{X}$ in the full-state system is independent of the state $S_{t}\in\mathcal{S}$ and the action $A_{t}\in\mathcal{A}$ (cf., Figure \ref{fig:diagram}). Assumption \ref{assump:s_dynamics} states that if the current state $S_t$, the action $A_t$ and $X_t$ (which is a state in the full-state system and an action in the reduced order model) are equal, then the probability of transitioning to any state $S_{t+1}$ for both systems is the same. These assumptions hold in various fields where the dynamics have a cascade structure. A concrete example is a quadrotor, where the pitch or roll angles influence the linear accelerations, but the latter do not affect the former (see Section \ref{sec_numerical_examples} for details).

 Note that the initial state distribution affects the trajectory and therefore the expected returns of the two models \eqref{eqn_optimal_reduced} and \eqref{eqn_high_order_controller_value}. Assumption \ref{assump:initial_dist} assumes that the distribution of the initial state is the same in both systems. Hence, it removes a source of discrepancy when evaluating the degradation and guarantees that the latter depends only on the difference in the dynamics of the systems.
\begin{assumption}\label{assump:initial_dist}  
Let $\mu_R(S_0)$ and $\mu_H(S_0)$ be the initial state distributions of the reduced-order and high-order models respectively. For all $S_0 \in \mathcal{S}$
    $\mu_R(S_0) = \mu_H(S_0)$.
\end{assumption}
We also make the following technical assumption regarding the transition probabilities of the reduced order model. 
\begin{assumption}\label{assump:lipschitz} The transition probability of the reduced model is $L$-Lipschitz continuous in $X$ in total variation norm. This is, for any $X_t, X_t^\prime \in \mathcal{X}$ it follows that 
\begin{align}
    \| \mathbb{P}_R(S_{t+1} \mid S_{t}, A_{t}, X_t^\prime) &- \mathbb{P}_R(S_{t+1} \mid S_{t}, A_{t}, X_t) \|_{TV} \nonumber \\
    &\leq L \| X_t^\prime - X_t \|.\label{LEX_t_X_t*}
\end{align}
\end{assumption}
%
In the case of dynamical systems with Gaussian noise, for instance, the total variation is bounded by a function proportional to the Euclidian distance between the two means~\cite{devroye2018total}. Since the mean is a deterministic function that depends on the system's dynamics for this assumption to hold it suffices that the dynamics are Lipschitz with respect to the state. This assumptions commonly holds e.g., Section \ref{sec_numerical_examples}. 


%
\section{Transfer guarantees}\label{sec_main_result}
In this section, we focus on establishing theoretical guarantees on the performance of a policy learned in the reduced order model when transferred to the full state dynamical system. As discussed earlier, the controller is in charge of tracking the reference $X_t^\star$ provided by the RL agent. It is intuitive that the better this reference is tracked the better the transferred policy will perform. As such, it is not unreasonable to require that the inner loop dynamics of $X$ and the controller to be stable. 
%
This is the subject of the following assumption.
\begin{assumption}\label{assump:contraction_property}
     Consider a closed loop system composed by the system with the transition $\mathbb{P}_{H}(X_{t+1} \mid X_{t}, U_{t})$ and the controller $U_t = K(X_t^\star, X_t)$. We assume that 
    there exists a positive definite matrix $P\in\mathbb{S}^{m\times m}_{++}$, and scalars $\alpha\in(0,1)$ and $\beta >0$ such that 
    \begin{align}\label{eqn_contraction_property_}
    \mathbb{E}\bigg[\|X_t - X_t^{\star} \|_P \bigg] 
    \leq \alpha &\mathbb{E}\bigg[\| X_{t-1} - X_{t-1}^{\star} \|_P\bigg]\nonumber \\& +\beta\mathbb{E}\bigg[\| X_{t}^{\star} - X_{t-1}^{\star}\|_P\bigg],
\end{align}
     where in the above equation \( \| \cdot \|_P \) denotes the norm induced by matrix $P$. Let $\lambda_{\max},\lambda_\min>0$ be the largest and smallest eigenvalues and define $\rho = \sqrt{\lambda_\max/\lambda_\min}$.
\end{assumption}
The above assumption relates to input-to-state stability (ISS) in expectation (see e.g.,~\cite{Khalil:1173048}). Indeed, notice that applying the previous expression recursively we have that 
\begin{align}\label{eqn_iss}
    \mathbb{E}\bigg[\|X_t - X_t^{\star} \|_P \bigg] 
    \leq \alpha^{t-1} &\mathbb{E}\bigg[\| X_{0} - X_{0}^{\star} \|_P\bigg] \\& +\beta\sum_{u=1}^t \alpha^{t-u}\mathbb{E}\bigg[\| X_{u}^{\star} - X_{u-1}^{\star}\|_P\bigg]. \nonumber
\end{align}
If we define a state $e_t=\mathbb{E}\left[\left\|X_t-X_t^\star\right\|\right]$ and an input $u_t =\mathbb{E}\left[\left\|X_t^\star-X_{t-1}^\star\right\|\right]$, the first term on the right hand side of the above equation is a class $\mathcal{K}\mathcal{L}$ function of $e_0$ and the second term is a class $\mathcal{K}$ function of $\left\|u\right\|_\infty$ (see e.g.,~\cite{Khalil:1173048}). 
Assumption \ref{assump:contraction_property} is, in general, not sufficient to provide guarantees in the performance degradation when transferring the policy. Indeed, the error between the state $X_t$ and the commanded action by the policy $X_t^\star$ depends on the variation of the commanded action. Since the $\mathcal{K}$ function is such that it goes to infinity when its argument goes to infinity, without guarantees on the boundedness of the variation $\mathbb{E}\bigg[\|X_{t+1}^\star - X_t^{\star} \|_P \bigg] $, the tracking error \eqref{eqn_iss} could be arbitrarily large. Thus, we introduce the following assumption which bounds the infinity norm of the variation.
\begin{assumption}\label{assump:contraction_property_policyR}
There exists $C>0$ such that
\begin{equation}
\mathbb{E}\bigg[\| X_t^\star - X_{t-1}^\star \|_P\bigg] \leq C,
\end{equation}
where $P$ is the matrix defined in Assumption \ref{assump:contraction_property}.
\end{assumption}
Note that, although the policy is trained using RL the above requirements can be promoted by penalizing large variations on the actions or even guaranteed if one expresses these requirements as constraints~\cite{chen2024probabilistic}. Under these assumptions we are now in conditions to establish a bound in the total variation difference between the transition probability of the reduced-order system and the high-order system with the inner loop controller. We define for simplicity
\begin{align}\label{eq:def_TV}
    TV(t+1)&:=   \sup_{S_t\in\mathcal{S},A_t\in \mathcal{A}, X_t^\star\in \mathcal{X}} \frac{1}{2}\sum_{S_{t+1} \in \mathcal{S}} \\&\left|\mathbb{P}_K(S_{t+1} \mid S_{t}, A_{t},X_t^\star)   -\mathbb{P}_R(S_{t+1} \mid S_{t}, A_{t},X_t^\star)\right|. \nonumber
\end{align} 
The previous quantity is the total variation between the transition probabilities of the reduced order and the full state system. It is important to point out that $TV(t)$ could be time-dependent. Indeed, as time progresses a stable inner controller should track better the commanded $X_t^\star$, resulting in dynamics that are closer to those of the reduced order-model. The next proposition formalizes this idea.
%
\begin{proposition}\label{proposition:---_stable} Under Assumption \ref{assump:x_dynamics}--\ref{assump:contraction_property_policyR}, 
the total variation defined in \eqref{eq:def_TV} satisfies
\begin{equation}
    TV(1)\leq \frac{L}{2} \mathbb{E}\left[\| X_{0} - X_{0}^{\star} \|\right],
    \end{equation}
and for all time $t\geq1$ it holds that
\begin{align}\label{eq:lemma_tv}
    TV(t+1)\leq \frac{L\rho}{2}\left(\alpha^{t}  \mathbb{E}\left[\| X_{0} - X_{0}^{\star} \|\right] + \beta C\frac{1-\alpha^{t}}{1 - \alpha} \right). 
\end{align}
\end{proposition}
\begin{proof}
    See Appendix \ref{sec_appendix_lemma_tv_x0x0*}.
\end{proof}

The main implication of the above result is that the difference between the dynamics of the reduced order model and the full state system is bounded for all $t \geq 0$. In particular, observe that the error due to the initial condition being different than the commanded state $X_0^\star$ decays exponentially. This is not surprising given the stability assumption made about the inner loop controller. Furthermore, the decay depends on the speed of the inner loop through the parameter $\alpha$ defined in Assumption \ref{assump:contraction_property}. The smaller $\alpha$ is, the smaller the bound in total variation. On the other hand, an error that depends on the bound of the variation of the signal ${X_t}^\star$ in consecutive time-steps is also present. Note that this error can only be made zero if the reference signal is constant $C=0$ or if $\beta =0$. The latter scenario requires the controller to be predictive. This is, to know at time $t$ the value of the next commanded state $X^\star_{t+1}$, which in turn depends on the RL policy and thus its value is only reveled at time $t+1$. Although such controller will, generally speaking, not be implementable, one could introduce complexity to the controller to better predict future RL actions. 

The key implication of the above result is that it allow us to bound the performance degradation of the optimal policy when transferred to the full state system. Note that the faster the controller is ($\alpha$ closer to zero), the smaller the error between the dynamics; hence, we should expect better transfer. This is the subject of the next theorem.

\begin{theorem}\label{theorem:2}
Under Assumption \ref{assump:reward_func}--\ref{assump:contraction_property_policyR}, the performance degradation by transfer can be bounded as    
\begin{align}\label{eqn_theorem2}
    (1-\gamma)\left|V_K^{\pi_R^\star} -V_R^\star\right| &\leq    BL\frac{\gamma^{2}}{(1 -\gamma\alpha)} \bigg( \frac{1}{(1-\gamma)}\beta C \\
    &+\left(1+\alpha (\rho -\gamma)\right)  \mathbb{E}\left[\| X_{0} - X_{0}^{\star} \|\right] \bigg). \nonumber 
\end{align}
%
\end{theorem}
\begin{proof}
   See Appendix \ref{sec_proof_theorem_2}.
\end{proof}

%
The previous theorem confirms the intuition derived earlier: if the initial state of the high-order system $X_0$ is equal to the initial commanded action $X_0^\star$ the loss by transfer is equal to zero in the exponentially stable case ($\beta =0$). This is not surprising since in the case of an exponentially stable controller the reference will be tracked without error if the initial error is zero. Thus, the dynamics of the reduced order and the high-order systems are the same. When there initial state does not correspond to the initial commanded action the effect of $\alpha$ can be observed. Indeed note that the bounds are monotonically increasing with $\alpha$. This confirms the intuition that a good controller (small $\alpha$) should achieve a better transfer than a poor controller ($\alpha \approx 1$). This is also the case under the ISS assumption. The term involving \(C\) and $\beta$ highlights the impact of the rate of change of the reference trajectory. A larger \(C\) (indicating greater changes between consecutive references) or $\beta$ (larger gain) leads to a larger bound. 
Lastly, it is important to notice that, as it is usual in problems with discounted infinite horizon, the more importance is given to the future (i.e., $\gamma \approx 1$), the larger the bound, whereas if $\gamma \approx 0$, the bound decreases. 



In the next section we focus on demonstrating the practical implications of the above theory in the problem of a quadrotor navigating to a desired destination.

\section{Experimental Results}\label{sec_numerical_examples}

In this section, we present numerical results evaluating the performance degradation when transferring a policy from a reduced-order model to a full state state system. We consider a simulated quadrotor navigation task. 
%
%
The horizontal and vertical positions of the quadrotor are given by $ [y,z]^\top \in [0,10]\times[0,10] $. The objective of the agent is to reach a fixed target at $[y_{\text{target}},z_{\text{target}}]^\top=[9,9]$. In the reduced-order model, the attitude of the quadrotor (pitch or roll) is considered an input, while in the high-order system, it is a state. In what follows we provide details of the two systems.


\subsection{Reduced-Order Model of a quadrotor}

The state for the reduced-order model consists of the horizontal and vertical positions and velocities $S = [ y, \dot{y}, z, \dot{z} ]^\top$. The action space is given by the Thrust and the attitude of the quadrotor
 $A= [T, \theta]^\top\in [-1, 1] \times \left[-\frac{\pi}{8}, \frac{\pi}{8}\right]$.
 %
%
%
We approximate the continuous-time dynamics of the agent by the following discrete-time state equations~\cite{beard2012quadrotor}
\begin{align}\label{dynamics}
\ddot{y}_{t+1} &= \frac{mg+T_t}{m} \sin{\theta_t}, 
&\ddot{z}_{t+1} = \frac{mg+T_t}{m} \cos{\theta_t} - g, \nonumber\\
\dot{y}_{t+1} &= \dot{y}_t + \ddot{y}_{t+1} \Delta t, &\dot{z}_{t+1} = \dot{z}_t + \ddot{z}_{t+1} \Delta t, \nonumber\\
y_{t+1} &= y_t + \dot{y}_{t+1} \Delta t, 
&z_{t+1} = z_t + \dot{z}_{t+1} \Delta t, 
\end{align}
with $m$ being the mass of the quadrotor, $g$ the earth gravitational constant, and $\Delta t$ the sampling time.  We set $m=1$ kg, $g=9.81 ms^{-2}$  and $\Delta t = 0.05$ s.


\subsection{High-Order Model of a quadrotor}

In addition to the state of the reduced order model, the high order model also includes the angle as part of the state $[S^\top ,\theta]^\top $. We consider first-order dynamics $\dot{\theta}=u$, where the input is given by a proportional controller $u=-K_p(\theta-\theta^\star)$ so that the state tracks the commanded angle. The exact discretization of the above dynamics yields 
\begin{equation}
    \theta_{t+1} = e^{-K_p\Delta t} \theta_t +(1- e^{-K_p\Delta t})\ \theta_t^\star.
\end{equation}
 Note that the gain of the controller determine the discrepancy between the reduced and higher order models
\begin{equation}
     \theta_{t+1}-\theta_{t+1}^\star = e^{-K_p\Delta t}\left( \theta_t-\theta_{t}^\star\right) +\left( \theta_t^\star-\theta_{t+1}^\star\right).
\end{equation}
 Along the lines of Assumption \ref{assump:contraction_property}, we have that $\beta =1$ and $\alpha = e^{-K_p\Delta t}$, which implies that the larger the gain the smaller $\alpha$. Our theoretical results predict that larger gains $K_p$ achieve better transfer guarantees as measured by the discrepancy in the value functions. 

\subsection{Training the agent in the reduced order system.}

\begin{table}
\centering
\caption{Training Hyperparameters for PPO}
\label{tab:training_parameters}
\begin{tabular}{lcc}
\toprule
\textbf{Hyperparameter} & \textbf{Value} & \textbf{Description} \\
\midrule
Learning rate & $3 \times 10^{-4}$ & Step size for the optimizer \\
Optimizer & Adam & Optimization algorithm \\
Number of episodes & $1,000,000$ & Total training episodes \\
Batch size & $64$ & Number of samples per  batch \\
Clip range $\epsilon$ & $0.2$ & PPO clipping parameter \\
\bottomrule
\end{tabular}
\end{table}

We start the section by defining the reward function. The main objective of the agent is to attain the target while avoiding collisions with the boundaries of the space. As such, we design the following reward to inform these goals
\begin{align}
    r(s_t,a_t,x_t) &= \mathbf{1}_{\text{target}}(s_t,a_t,x_t) - \frac{\| p_t - p_{\text{target}} \|}{d_{\text{max}}} \nonumber \\&- C \mathbf{1}_{\text{boundary}}(s_t,a_t,x_t), \label{eqn_reward}
\end{align}
where $\mathbf{1}_{\text{target}}$ denotes the indicator function taking the value if $\|p_t - p_{\text{target}}\|_\infty \leq 0.05$ and  zero otherwise and $\mathbf{1}_{\text{boundary}}$ is the indicator function taking the value one if $p_t \notin [0,10] \times [0,10]$ and zero otherwise.
%
%
%
The second term on the right-hand side of  \eqref{eqn_reward} represents the normalized distance between the agent and the target, with \( d_{\text{max}}=10\sqrt{2} \) being the maximum possible distance within the environment. We set the value of the penalty for violating the environement's boundaries to \( C = 5 \times 10^3 \). The discount factor \( \gamma \) is set to 0.995.



We train the agent in the reduced order model using Proximal Policy Optimization (PPO)~\cite{Schulman2017ProximalPO}. The training hyperparameters are listed in Table~\ref{tab:training_parameters}.

\subsection{Policy Transfer and Evaluation}

\begin{figure}
    \centering
    \includegraphics[width=\linewidth]{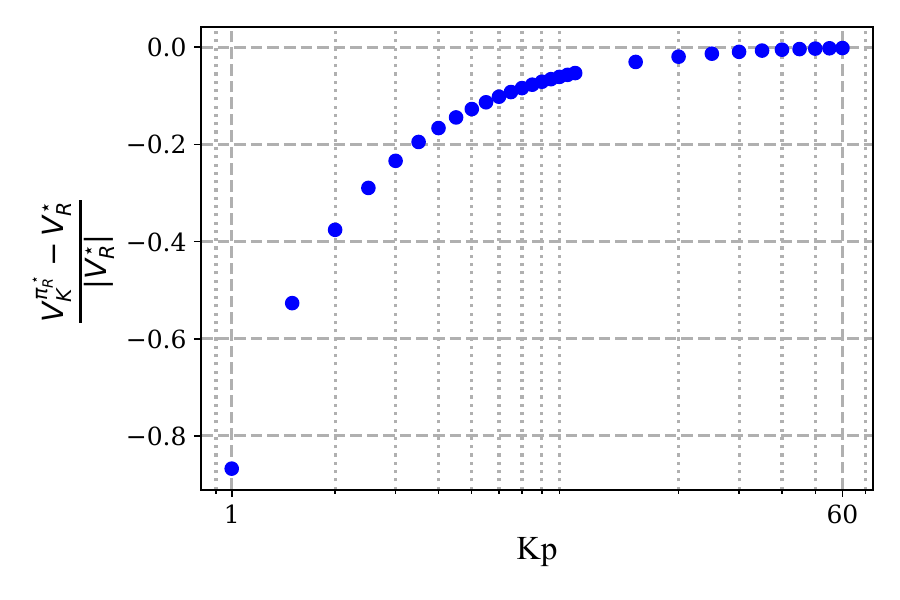}
    \caption{ Illustration of the relative average differences in expected discounted returns between the full-state system~\eqref{eqn_high_order_controller_value} and the reduced-order model~\eqref{eqn_optimal_reduced} for various values of \( K_p \). These are averaged over 100 experiments with randomly initialized states \( S \). In accordance with Theorem~\ref{theorem:2}, we observe that increasing \( K_p \) results in less performance degradation.}
    \label{fig:vk_vr}
\end{figure}
\begin{figure}
    \centering
    \includegraphics[width=\linewidth]{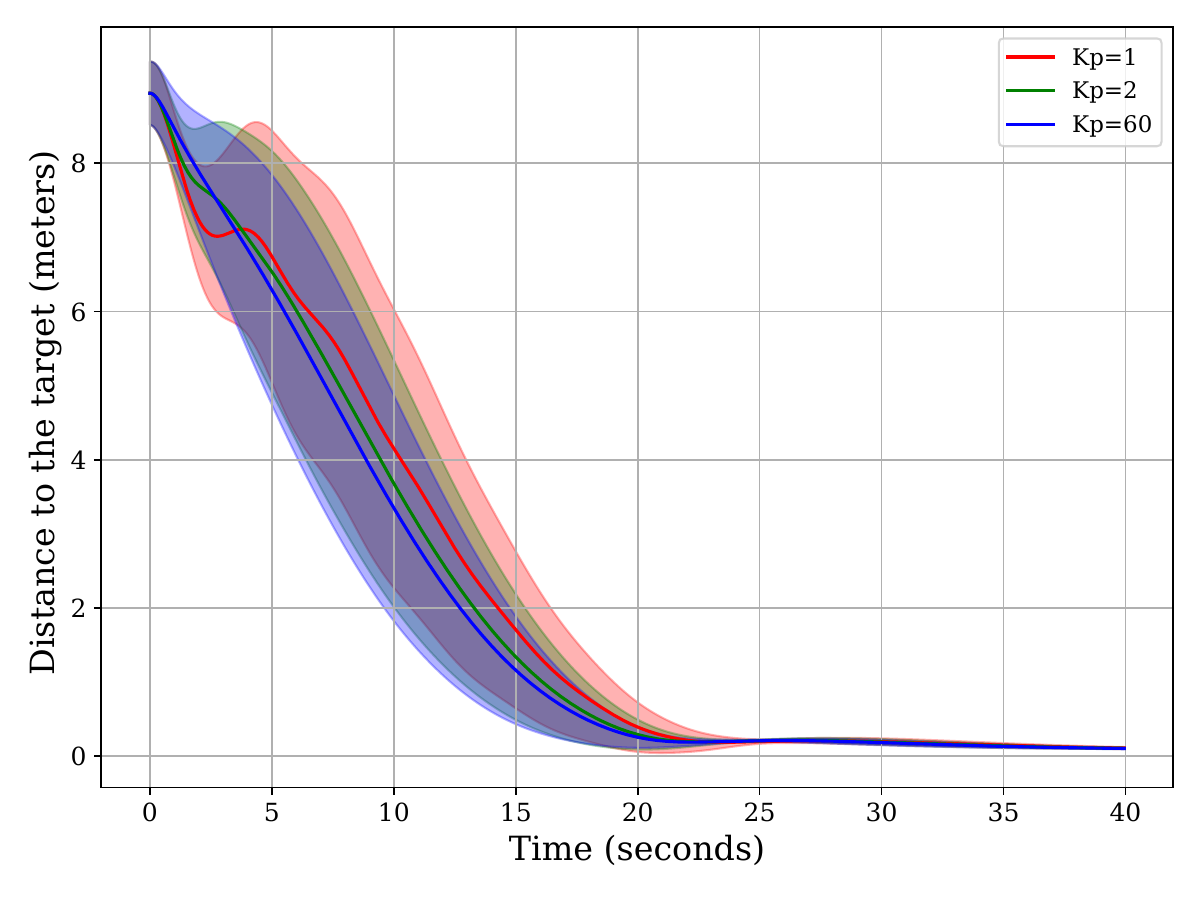} 
    \caption{ Mean and standard deviation of the distance to the target over 100 iterations for proportional gains $K_p$  set to 1, 2 and 60.}  
    \label{fig:distance}
\end{figure}
  
We assess the performance degradation of the transferred policy by examining the differences in expected discounted returns between the full state system \eqref{eqn_high_order_controller_value} and the reduced-order model \eqref{eqn_optimal_reduced} for different values of \( K_p \). Figure~\ref{fig:vk_vr} shows the mean differences in expected discounted rewards, calculated by averaging the returns over 100 iterations with random initialization of the state \( S \).
Consistent with Theorem~\ref{theorem:2}, we observe that a larger \( K_p \) leads to a smaller degradation of the performance. The reason for the difference in performance is due to the distance to the target that each system attains. Indeed, the reward function \eqref{eqn_reward} penalizes the distance to the target. As it can be observed in Figure \ref{fig:distance}, the larger the gain the faster the convergence to the target.

\begin{figure}
    \centering
    \includegraphics[width=\linewidth]{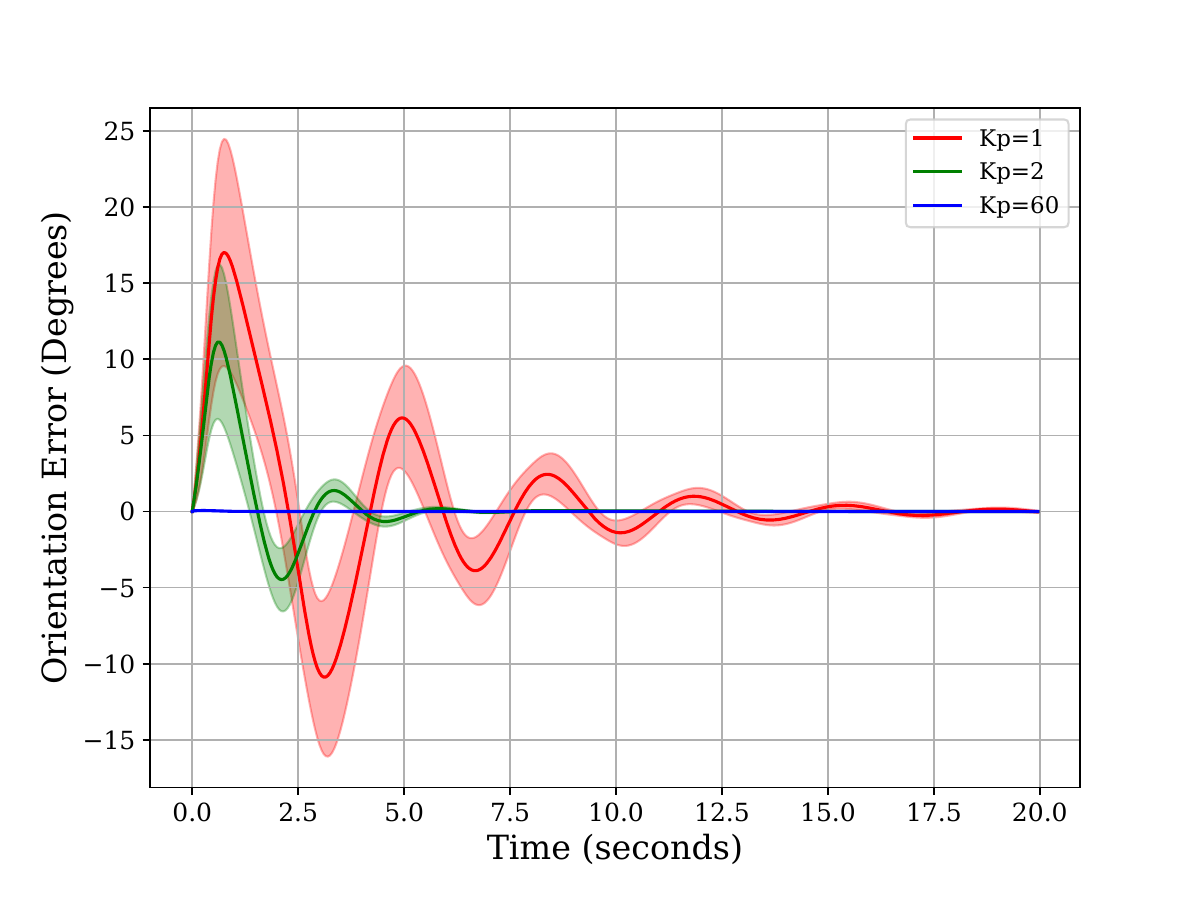} 
    \caption{Mean and standard deviation of the orientation error computed over 100 iterations with random initial states \( S \). The plot illustrates that as \( K_p \) increases, the orientation error approaches zero. Thus reducing the total variation between the two transitions. This result aligns with Proposition~\ref{proposition:---_stable}. }
    \label{fig:Error_theta}
\end{figure}

Figure~\ref{fig:Error_theta} shows the mean and standard deviation of the orientation error, computed over 100 iterations with random initial states \( S \). The plot shows as \( K_p \) increases, the orientation error approaches zero. This outcome aligns with Proposition~\ref{proposition:---_stable}, which states that a larger \( K_p \) (smaller $\alpha$) reduces the discrepancy between the transition probabilities of the full-state and reduced-order models aligning their dynamics since both depend on \( \theta \). Thus improving the transfer to the full-state system.

\begin{figure}
    \centering
    \includegraphics[width=\linewidth]{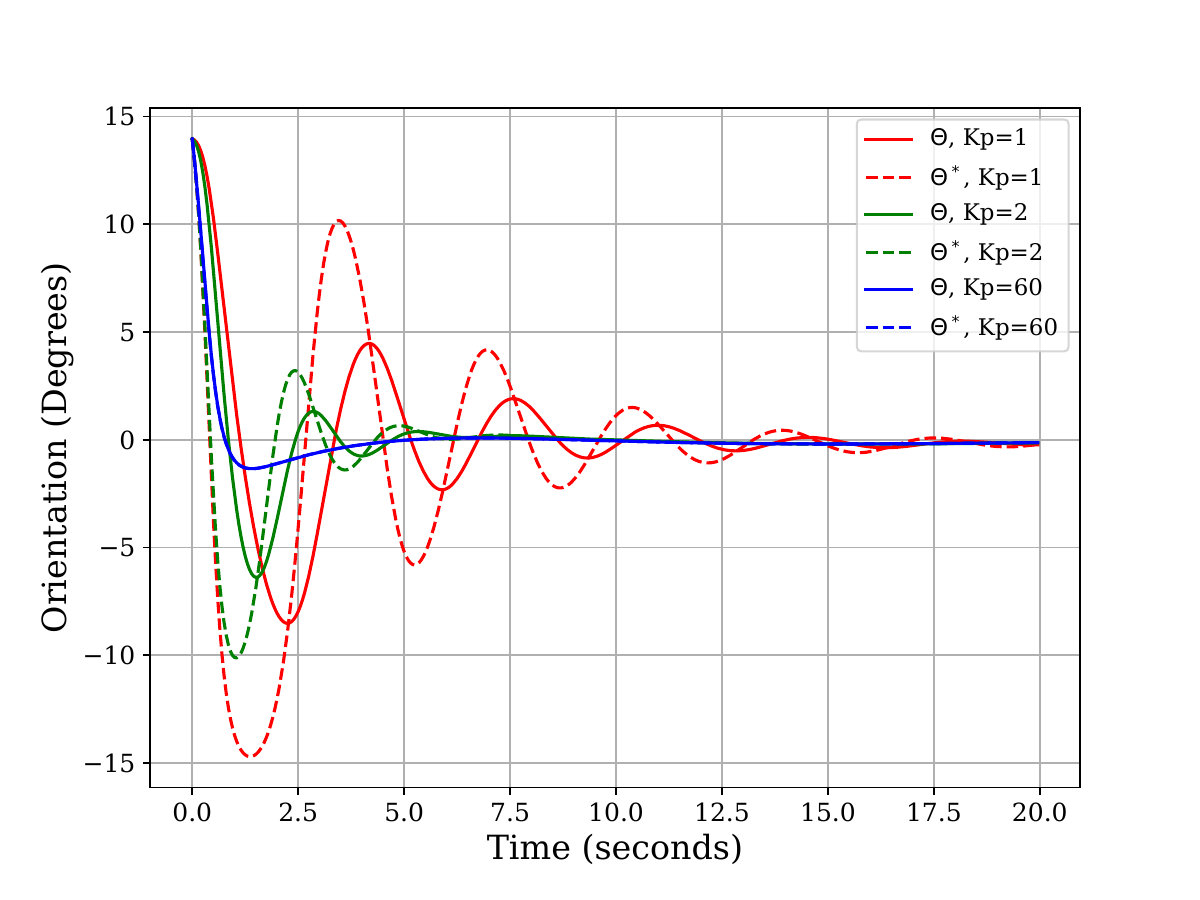} 
    \caption{Comparison of the orientation trajectories of the full-state and reduced-order models for different values of \( K_p \). Similar to Figure \ref{fig:Error_theta} we observe that the larger $K_p$ the more similar the trajectories between $\theta$ and the reference $\theta^\star$. Furthermore, we observe that the larger $K_p$ the less variations there are on $\theta^\star$. According to Theorem \ref{theorem:2} a large variation on the reference results in worse transfer which is consistent with this experiment.}
    \label{fig:thetas}
\end{figure}

Furthermore, Figure~\ref{fig:thetas} compares the orientation trajectories of the full-state and reduced-order models. Note that with a small value of $K_p$, the variation in the orientation of the reduced-order model (dashed lines) is larger. This is, there is more commanded effort. Since the bound in Theorem \ref{theorem:2} depends on this variation, the larger the bound in this variation the more degradation in the transfer one can expect. This effect added to the larger value of $\alpha$ for smaller gains $K_p$ explains the degradation curve in Figure \ref{fig:vk_vr}.




\section{Conclusion}
In this work we considered the transfer of an optimal policy learned in a reduced order model into a full state system for a class of cascade systems. Our theoretical guarantees establish that the performance degradation depends on the stability properties of the inner loop controller. In particular, the better the commanded signal by the RL policy can be tracked by the state of the high-order system, the smaller the loss in performance. We verified our theoretical findings with the example of a quadrotor, where its attitude corresponds to the inner state that is an action for the reduced order model. 

\section*{Acknowledgement}
This work was sponsored by the Office of Naval Research
(ONR), under contract number N00014-23-1-2377.

\appendix
\section{Appendix}\label{sec_appendix}

\subsection{Proof  of Proposition \ref{proposition:---_stable} }\label{sec_appendix_lemma_tv_x0x0*}
Marginalizing the joint probability of $S_{t+1}$ and $X_{t}$ and using Bayes rule it follows that 
\begin{align}
    \mathbb{P}_K( S_{t+1} \mid &S_{t}, A_{t}, X_{t}^\star)  \label{eqn_marginalization}\\ &= \sum_{X_{t}\in\mathcal{X}} \mathbb{P}_K( S_{t+1} \mid S_{t}, A_{t},X_{t}, X_{t}^\star) \mathbb{P}_K(X_{t}) \nonumber\\
    &= \sum_{X_{t}\in\mathcal{X}} \mathbb{P}_H( S_{t+1} \mid S_{t}, A_{t},X_{t}, X_{t}^\star) \mathbb{P}_K(X_{t}). \nonumber
\end{align}
The second equality holds since the impact of the controller is only on defining $X_t$ and not on the transitions of $S_t$ given $X_t$. Furthermore, using that the transitions of the reduced and high-order systems are related by Assumption \ref{assump:s_dynamics} and combining the above expression with \eqref{eqn_marginalization} it follows that 
\begin{align}
    \mathbb{P}_K( S_{t+1} \mid &S_{t}, A_{t}, X_{t}^\star)\label{eqn_bound_PK}  \\ &= \sum_{X_{t}\in\mathcal{X}} \mathbb{P}_R( S_{t+1} \mid S_{t}, A_{t},X_{t}) \mathbb{P}_K(X_{t}). \nonumber
\end{align}
On the other hand, since the transitions of the reduce order model are independent of the variable $X_t$, it follows that 
\begin{align}
    \mathbb{P}_R( S_{t+1} \mid &S_{t}, A_{t}, X_{t}^\star) \label{eqn_expanded_PR} \\ &= \sum_{X_{t}\in\mathcal{X}} \mathbb{P}_R( S_{t+1} \mid S_{t}, A_{t},X^\star_{t}) \mathbb{P}_K(X_{t}).\nonumber
\end{align}
Combining \eqref{eqn_bound_PK} and \eqref{eqn_expanded_PR} one can write the difference of transition probabilities in both MDPs as 
\begin{align}
   \mathbb{P}_K( S_{t+1} \mid S_{t}, A_{t}, X_{t}^\star) &-  \mathbb{P}_R( S_{t+1} \mid S_{t}, A_{t}, X_{t}^\star) \\ 
   = \sum_{X_{t}\in\mathcal{X}} \Big(\mathbb{P}_R( S_{t+1} \mid &S_{t}, A_{t},X_{t}) \nonumber\\ &-\mathbb{P}_R( S_{t+1} \mid S_{t}, A_{t},X^\star_{t}) \Big)\mathbb{P}_K(X_{t}). \nonumber
\end{align}
From the triangle inequality and the definition of total variation (see \eqref{eq:def_TV}) it follows that 

\begin{align}
2 TV(t+1)
\leq \sum_{S_{t+1}\in\mathcal{S}} &
    \sum_{X_{t}\in\mathcal{X}} \Big|\mathbb{P}_R( S_{t+1} \mid S_{t}, A_{t},X_{t}) \\ &-\mathbb{P}_R( S_{t+1} \mid S_{t}, A_{t},X^\star_{t}) \Big|\mathbb{P}_K(X_{t}). \nonumber
\end{align}
In the above expression, we have also used that $\mathbb{P}_K(X_t)$ is non-negative. We can further upper bound the above expression by using that the transition probabilities of the reduced order model are Lipschitz with respect to $X$ (Assumption \ref{assump:lipschitz}). Doing so, the above expression reduces to
\begin{align}\label{eqn_TV_P}
&2 TV(t+1)
\nonumber \\
&\leq L \sum_{X_{t}\in\mathcal{X}} \|X_{t}-X^\star_{t}\|\mathbb{P}_K(X_{t}) =L \mathbb{E} \left[\|X_{t}-X^\star_{t}\| \mid X^\star_t\right],
\end{align}
where the equality follows directly from the definition of conditional expectation. Observe that for the above expression to hold, $X_t$ needs to be independent of $X_t^\star$.  

For any positive definite matrix \( P \) with eigenvalues \( \lambda_{\text{min}} \) and \( \lambda_{\text{max}} \), the following inequality holds (see e.g., \cite{horn2013matrix})
\begin{align}\label{eq_norm_relation}
\sqrt{\lambda_{\text{min}}} \| X_{t} - X_{t}^\star \| &\leq \| X_{t} - X_{t}^\star \|_P \nonumber\\&\leq \sqrt{\lambda_{\text{max}}} \| X_{t} - X_{t}^\star \|.
\end{align}


 Using \eqref{eq_norm_relation} and applying \eqref{eqn_contraction_property_} recursively, it follows that
\begin{align}\label{bound_E_Xt_X*t_---_stbl}
    &\mathbb{E} \bigg[\|  X_{t} - X_{t}^{\star} \| \bigg] \leq \frac{1}{\sqrt{\lambda_{\text{min}}}}\mathbb{E}\bigg[\| X_{t} - X_{t}^{\star} \|_P \bigg] \\&\leq \frac{1}{\sqrt{\lambda_{\text{min}}}} \alpha^{t} \mathbb{E}\bigg[\| X_{0} - X_{0}^{\star} \|_P\bigg]
    +\sum_{k=0}^{t-1}\alpha^k\beta
    \mathbb{E}\bigg[\| X_{t}^\star - X_{t-1}^{\star} \|_P\bigg]. \nonumber  
    \end{align}
Substituting the bound in Assumption \eqref{assump:contraction_property_policyR} into \eqref{bound_E_Xt_X*t_---_stbl} and writing the geometric sum compactly yields 

\begin{align}\label{bound_E_Xt_X*t_---_stbl_}
    &\mathbb{E} \bigg[\|  X_{t} - X_{t}^{\star} \| \bigg] \leq  \nonumber \\ &\frac{\sqrt{\lambda_{\text{max}}}}{\sqrt{\lambda_{\text{min}}}}\left(\alpha^{t}  \mathbb{E}\left[\| X_{0} - X_{0}^{\star} \|\right] + \beta C\frac{1-\alpha^{t})}{1 - \alpha} \right).
\end{align}
Substituting the bound \eqref{bound_E_Xt_X*t_---_stbl_} into \eqref{eqn_TV_P} completes the proof of the result. \hfill \QED

\subsection{Proof of Theorem \ref{theorem:2}.}\label{sec_proof_theorem_2}

To support the proofs of Theorem \ref{theorem:2}, it is essential to bound the discrepancy between the trajectory distributions of the high-order system with the inner-loop controller and the reduced-order MDP. This discrepancy is measured by \(\Delta P{(t)}\), defined as
%
%
\begin{align}\label{delatP}
\Delta P{(t)} := \sum_{(s,a)_{0:t}} \left| \mathbb{P}_K((s,a)_{0:t}) - \mathbb{P}_R((s,a)_{0:t}) \right|.
\end{align}

\begin{lemma}\label{bound_for_sum|PK_Pr|_}
Under Assumptions \ref{assump:initial_dist}--\ref{assump:contraction_property_policyR}, the total variation distance between the trajectory distributions over the time horizon \(t\) satisfies $\Delta P(0)=0$ and for $t\geq1$ it holds that
\begin{align}
    &\Delta P(t) \leq L \| X_{0} - X_{0}^{\star} \| + L\rho \frac{\beta C}{1 - \alpha}(t-1)  \\ &+  L\rho\alpha \frac{1-\alpha^{t-1}}{1-\alpha} \left(\mathbb{E}\left[\| X_{0} - X_{0}^{\star} \|\right] - \frac{\beta C}{1 - \alpha} \right).  \nonumber 
    \end{align}
%
\end{lemma}
\begin{proof}
See Appendix \ref{sec_for_sum|PK_Pr|}.
\end{proof}

We begin by computing the difference between the expected cumulative returns associated with each MDP. Using the definition of expected cumulative return (see~\eqref{eqn_optimal_reduced} and \eqref{eqn_high_order_controller_value}) it follows that
\begin{align}\label{vk_vr}
    V_K^{\pi^\star_R} - V_R^\star &= \mathbb{E}^{\pi}_K \bigg[ \sum_{t=0}^{\infty} \gamma^t r(s_t,a_t,x_t)\bigg] \\& - \mathbb{E}^{\pi}_R \bigg[ \sum_{t=0}^{\infty} \gamma^t r(s_t,a_t,x_t)\bigg].  \nonumber
\end{align}

Writing the expectation as a sum and exchanging the order of the sums, we have that
\begin{align}\label{expt_definition}
    V_K^{\pi_R^\star} &=\sum_{t=0}^{\infty}\sum_{(s,a,x)_{0:\infty}}  \gamma^t r(s_t,a_t,x_t)\mathbb{P}_K^{\pi_R^\star}((s,a,x)_{0:\infty} \ldots ),
\end{align}
where in the above notation we use $(s,a,x)_{0:\infty}$ to denote all states an actions. Since the rewards are bounded (Assumption \ref{assump:reward_func}), the sums in \eqref{vk_vr} are bounded. Thus, the Tonelli-Fubini Theorem (see e.g., ~\cite{rudin1987real}) justifies the exchange of sum over time and expectation. 
We further claim that
\begin{align}\label{eqn_variables_until_t}
    V_K^{\pi_R^\star} &=\sum_{t=0}^{\infty}\sum_{(s,a,x)_{0:t}}  \gamma^t r(s_t,a_t,x_t)\mathbb{P}_K^{\pi_R^\star}((s,a,x)_{0:t} \ldots )
\end{align}
and analogously that 
\begin{align}\label{eqn_variables_until_t2}
    V_R^{\star} &=\sum_{t=0}^{\infty}\sum_{(s,a,x)_{0:t}}  \gamma^t r(s_t,a_t,x_t)\mathbb{P}_R^{\star}((s,a,x)_{0:t} \ldots ).
\end{align}
We defer the proof of these claims to the end of the proof.

Subtracting \eqref{eqn_variables_until_t2} to \eqref{eqn_variables_until_t} taking the absolute value on both sides of the above equation, using the triangle inequality and the fact that $\left|r(s,a)\right|\leq B$ (Assumption \ref{assump:reward_func}) it follows that 
\begin{align}\label{eq:Bound_|Vk_Vr|}
    &\left|V_K^{\pi_R^\star} -V_R^\star\right|\leq    B\sum_{t=0}^{\infty}\sum_{(s,a)_{0:t}}  \gamma^t \Delta P(t)
\end{align}
where we have also used the definition of $\Delta P(t)$ (see \eqref{delatP}). Then, by virtue of Lemma \ref{bound_for_sum|PK_Pr|_} it follows that  
\begin{align}
&\left|V_K^{\pi_R^\star} -V_R^\star\right|\leq  BL \sum_{t=1}^{\infty}\gamma^t  \left(\mathbb{E}\left[\| X_{0} - X_{0}^{\star} \|\right]+\rho \frac{\beta C}{1 - \alpha} (t-1)\right)  \nonumber \\ &+  B L\rho\alpha  \left(\mathbb{E}\left[\| X_{0} - X_{0}^{\star} \|\right] - \frac{\beta C}{1 - \alpha} \right)\sum_{t=1}^{\infty}\gamma^t\frac{1-\alpha^{t-1}}{1-\alpha}.   \nonumber \\ 
\end{align}

From the convergence of the following geometric series (see e.g.,~\cite{apostol1974mathematical})

\begin{align}\label{series1}
    \sum_{t=1}^{\infty}\gamma^t =  \frac{\gamma^{2}}{1 -\gamma},\;\sum_{t=1}^{\infty}\gamma^t (t-1) =\frac{\gamma^{2}}{(1 - \gamma)^{2}}
\end{align}

\begin{align}\label{series2}
\sum_{t=1}^{\infty}\gamma^t\frac{1-\alpha^{t-1}}{1-\alpha} = \frac{\gamma}{(1- \gamma)(1 -\gamma\alpha)}\
\end{align}
it follows that
\begin{align}
    &\left|V_K -V_R\right|\leq  BL \frac{\gamma^{2}}{1 -\gamma}  \left(\| X_{0} - X_{0}^{\star} \| +\rho \frac{\beta C}{(1 - \alpha)(1-\gamma)}\right) \nonumber \\ &+   BL\rho\alpha  \left(\mathbb{E}\left[\| X_{0} - X_{0}^{\star} \|\right] - \frac{\beta C}{1 - \alpha} \right)\frac{\gamma^{2}}{(1- \gamma)(1 -\gamma\alpha)}.
\end{align}
Re-arranging the above expressions yields \eqref{eqn_theorem2}. To complete the proof we are left to prove \eqref{eqn_variables_until_t} and \eqref{eqn_variables_until_t2}. We will prove the latter. The former follows the same steps. Split the sum into the sum until time $t$ and another term after $t+1$ 
\begin{align}\label{eq:0_to_t_infinity}
    &\sum_{(s,a)_{0:\infty}}  \gamma^t r(s_t,a_t,x_t) \mathbb{P}_R^\star((s,a,x)_{0:\infty} = \\
    &\sum_{(s,a,x)_{0:t}} \gamma^t r(s_t,a_t) \mathbb{P}_R^\star(s,a,x)_{0:t}\\ &\nonumber\sum_{(s,a,x)_{t+1:\infty}} \mathbb{P}_R^\star\left( (s,a,x)_{t+1: \infty} \mid (s,a,x)_{0:t}\right).\nonumber
\end{align}

Since $\mathbb{P}\left( (s,a)_{t+1: \infty} \mid (s,a)_{0:t}\right)$ is a probability the rightmost sum in the above equation is equals to one. Therefore, the above equation reduces to \eqref{eqn_variables_until_t}. \hfill \QED

\subsection{Proof of Lemma \ref{bound_for_sum|PK_Pr|_}} \label{sec_for_sum|PK_Pr|}

Applying Bayes' rule, using that the policy depends only on the state $s_t$ and the Markov property of the transitions it follows that 
\begin{align}
&\mathbb{P}_R^\star((s,a,x)_{0:t})= \\&\mathbb{P}_R^\star((s,a,x)_{0:t-1})  \mathbb{P}_R(s_t \mid s_{t-1}, a_{t-1},x_{t-1})  \pi^\star_R(a_t,x_t \mid s_t). \nonumber
\end{align} 
Analogously it follows that 
\begin{align}
&\mathbb{P}_K^{\pi_R^\star}((s,a,x)_{0:t})= \\&\mathbb{P}_K^{\pi_R^\star}((s,a,x)_{0:t-1})  \mathbb{P}_H(s_t \mid s_{t-1}, a_{t-1},x_{t-1})  \pi^\star_R(a_t,x_t \mid s_t). \nonumber
\end{align}
Using these expressions, we can write \(\Delta P(t)\) (see \eqref{deltaP}) as
\begin{align}
    &\Delta P{(t)} = \sum_{a_t,x_t}\pi(a_t,x_t\mid s_t)  \sum_{(s,a,x)_{0:t-1}, s_t} \\& \left| \mathbb{P}_K^{\pi_R^\star}((s,a,x)_{0:{t-1}}) \mathbb{P}_H(s_t \mid s_{t-1}, a_{t-1},x_{t-1}) \right. \nonumber \\
    &  - \mathbb{P}_R^\star((s,a,x)_{0:{t-1}}) \mathbb{P}_R(s_t \mid s_{t-1}, a_{t-1},x_{t-1}) \bigg|\nonumber
\end{align}
Since $\pi(a_t,x_t\mid s_t)$ is a probability distribution, the leftmost sum in the above equation equals one. Therefore, by adding and subtracting the term
$
\mathbb{P}_R(s_t \mid s_{t-1}, a_{t-1},x_{t-1}) \mathbb{P}_K^{\pi_R^\star}((s,a,x)_{0:t-1})$ to each term in the above sum yields
\begin{align}\label{eqn_auxiliary_lemma1}
    &\Delta P{(t)} =   \sum_{(s,a,x)_{0:t-1}, s_t} \mathbb{P}_K^{\pi_R^\star}((s,a,x)_{0:{t-1}}) \left|\mathbb{P}_H-\mathbb{P}_R\right|\\
    &  + \left|\mathbb{P}_K^{\pi_R^\star}((s,a,x)_{0:{t-1}})-\mathbb{P}_R^\star((s,a,x)_{0:{t-1}})\right| \mathbb{P}_R,\nonumber
\end{align}
where in the above expression we have omitted the variables of the transition probabilities for simplicity in the notation. Using the definition of the total variation in \eqref{eq:def_TV} we can upper bound the first term of the above sum as 
\begin{align}\label{eqn_auxiliary_lemma12}
&\sum_{(s,a,x)_{0:t-1}, s_t} \mathbb{P}_K^{\pi_R^\star}((s,a,x)_{0:{t-1}}) \left|\mathbb{P}_H-\mathbb{P}_R\right| \leq \\
 &2\sum_{(s,a,x)_{0:t-1}} \mathbb{P}_K^{\pi_R^\star}((s,a,x)_{0:{t-1}}) TV(t-1) =2TV(t-1), \nonumber
\end{align}
where the equality follows from the fact that $ \mathbb{P}_K^{\pi_R^\star}((s,a,x)_{0:{t-1}})$ is a probability distribution.
Using the fact that $\mathbb{P}_R$ is also a probability distribution, we can write the second term in \eqref{eqn_auxiliary_lemma1} as
\begin{align}\label{eqn_auxiliary_lemma13}
    \sum_{(s,a,x)_{0:t-1}, s_t}&
\left|\mathbb{P}_K^{\pi_R^\star}((s,a,x)_{0:{t-1}})-\mathbb{P}_R^\star((s,a,x)_{0:{t-1}})\right| \mathbb{P}_R \nonumber \\ 
&= \Delta P(t-1).
\end{align}
Substituting \eqref{eqn_auxiliary_lemma12} and \eqref{eqn_auxiliary_lemma12} into \eqref{eqn_auxiliary_lemma1} reduces to
\begin{align}\label{eq:deltaP_TV}
    \Delta P(t) &\leq 2 TV(t)+ \Delta P(t-1)\leq  2 \sum_{u=1}^t TV(u). 
\end{align}
Where the second inequality follows from applying the left-most inequality recursively and using the fact that $\Delta P(0) = 0$. The latter holds since by Assumption \ref{assump:initial_dist} the initial distribution of the state is the same for both systems. The proof is then completed by replacing the total variation by the bounds in Proposition \ref{proposition:---_stable}.

The proof of the result is completed by rearranging the terms and simplifying the sums. \hfill \QED

\bibliographystyle{ieeetr}


\end{document}